\documentclass[10pt,twocolumn,letterpaper]{article}
\usepackage[accsupp]{axessibility}
\usepackage{iccv}
\usepackage{times}
\usepackage{epsfig}
\usepackage{graphicx}
\usepackage{amsmath}
\usepackage{amssymb}
\usepackage{booktabs}
\usepackage{multirow}

\usepackage[pagebackref=true,breaklinks=true,letterpaper=true,colorlinks,bookmarks=false]{hyperref}

\iccvfinalcopy 

\def\iccvPaperID{7278} 

\ificcvfinal\fi

\begin{document}
	
	\title{Supervised Homography Learning with Realistic Dataset Generation}
	\author{
		Hai Jiang\textsuperscript{\rm1,\rm2,\thanks{Equal contribution}}, Haipeng Li\textsuperscript{\rm3,\rm2,\footnotemark[1]}, Songchen Han\textsuperscript{\rm1,\footnotemark[2]}, Haoqiang Fan\textsuperscript{\rm2}, Bing Zeng\textsuperscript{\rm3}, Shuaicheng Liu\textsuperscript{\rm3,\rm2,\thanks{Corresponding authors}}\\
		\textsuperscript{\rm1}Sichuan University \textsuperscript{\rm2}Megvii Technology \\
		\textsuperscript{\rm3}University of Electronic Science and Technology of China \\
		\tt\small\{jianghai1@stu., hansongchen@\}scu.edu.cn, \tt\small\{jianghai,lihaipeng,fhq,liushuaicheng\}@megvii.com \\ \tt\small\{lihaipeng@std.,eezeng@,liushuaicheng@\}uestc.edu.cn
	}

	\maketitle
	\ificcvfinal\thispagestyle{empty}\fi
	
	\begin{abstract}
		In this paper, we propose an iterative framework, which consists of two phases: a generation phase and a training phase, to generate realistic training data and yield a supervised homography network. In the generation phase, given an unlabeled image pair, we utilize the pre-estimated dominant plane masks and homography of the pair, along with another sampled homography that serves as ground truth to generate a new labeled training pair with realistic motion. In the training phase, the generated data is used to train the supervised homography network, in which the training data is refined via a content consistency module and a quality assessment module. Once an iteration is finished, the trained network is used in the next data generation phase to update the pre-estimated homography. Through such an iterative strategy, the quality of the dataset and the performance of the network can be gradually and simultaneously improved. Experimental results show that our method achieves state-of-the-art performance and existing supervised methods can be also improved based on the generated dataset. Code and dataset are available at https://github.com/JianghaiSCU/RealSH.
	\end{abstract}
	
	\section{Introduction}
	\label{sec:intro} 
	\begin{figure}[!t]
		\centering
		\includegraphics[width=\linewidth]{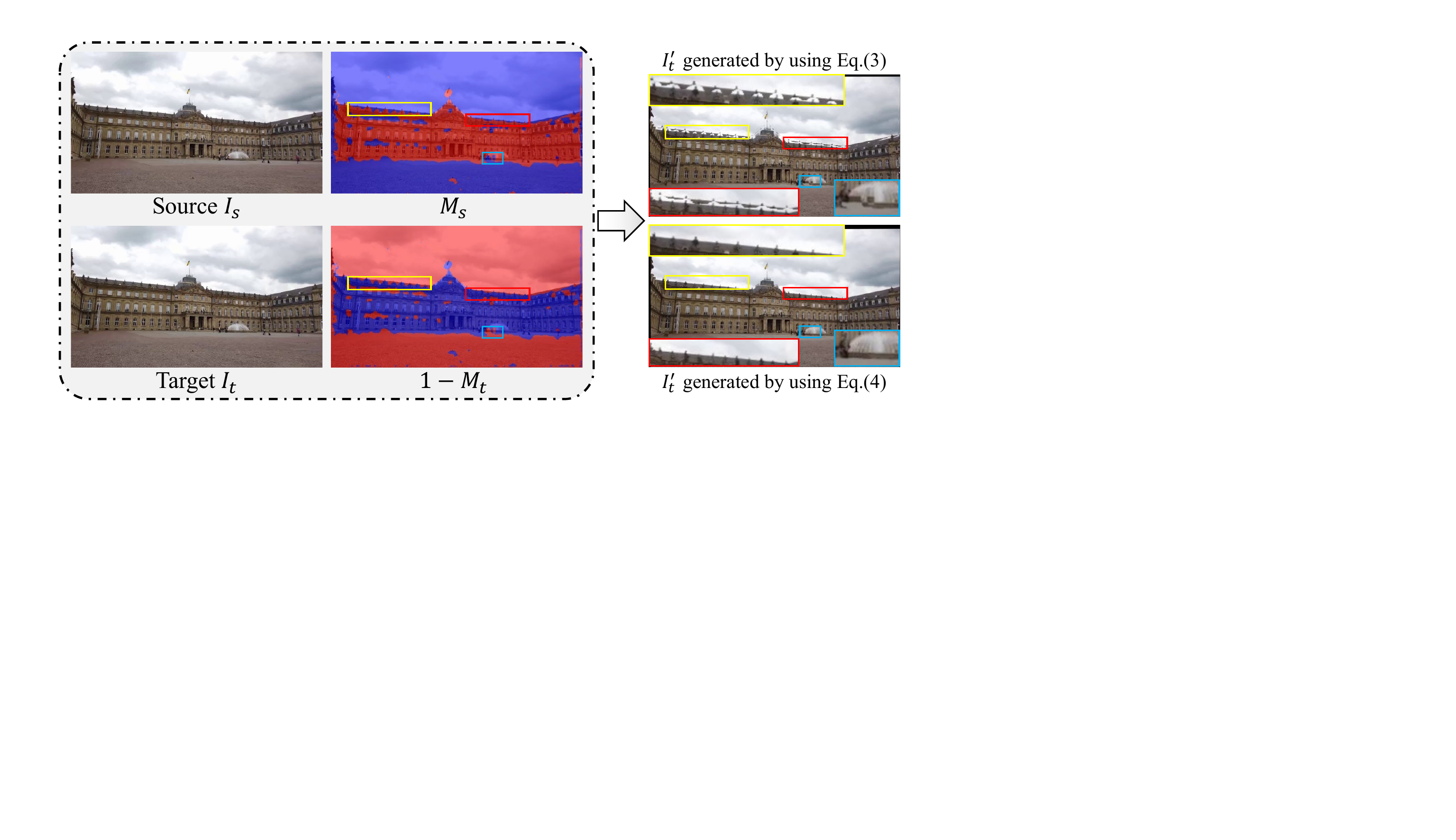}
		\caption{The first row shows the target images generated by the previous strategy~\cite{supervised2016} and our proposed method, visualized by superimposing the source image warped by GT homography on the generated target image, where the misaligned pixels are represented as colored ghosts. As highlighted in yellow boxes, the dominant plane is fully aligned by the GT homography, proving that both our method and the previous strategy can satisfy the label criteria. In contrast, only our method can maintain realistic motion between the two images thus satisfying the realism criteria, as shown in the red boxes, the car is a moving object that cannot be aligned by a homography.}
		\label{fig:teaser}
	\end{figure}
	
	Homography estimation is a fundamental task in computer vision that has been widely used for high-dynamic range imaging~\cite{HDR1,HDR2,HDR3}, image stitching~\cite{stitching2,stitching3}, and video stabilization~\cite{stabilization1,stabilization2}. Traditional methods typically use feature extraction and matching methods~\cite{sift,surf} with outlier suppression~\cite{ransac,magsac} to obtain feature matches of two images and subsequently solve direct linear transformation (DLT)~\cite{DLT} to obtain the homography matrix. However, these methods highly rely on the extracted matching keypoints and may crash in challenging scenes that lack sufficient high-quality feature matches. With the rise of deep learning, such problems have been partially solved, learning-based methods~\cite{CA-TPAMI,BasesHomo_Pami,LBHomo} take a pair of images as input and directly output the corresponding homography, thus are more robust than traditional methods due to their keypoint-free estimation strategy. The learning-based methods can be divided into two categories: supervised and unsupervised. At the moment, benefiting from the label-free characteristic, unsupervised methods can be trained on enormous amounts of real-world data, delivering better performance and generalization capability than supervised ones. We find that the lack of qualified training data is one of the main barriers to the development of supervised methods. Previously supervised dataset generation strategy~\cite{supervised2016} synthesizes an image pair by using pre-defined ground truth (GT) homography to warp a single image, such strategy considers the whole image as a single plane, neglecting parallax and foreground motion in the real world. As shown in Fig.~\ref{fig:teaser}(a), the two images are fully aligned using the GT homography, but the car in the red box should be a moving object that cannot be aligned by a homography.
	
	In this work, we propose an iterative framework, which is designed to generate image pairs that satisfy both label and realism criteria~\cite{RealFlow} for supervised homography learning and learns a state-of-the-art homography estimation network with the generated dataset. Specifically, given an unlabeled source and target image ($I_s$ and $I_t$) captured in real-world scenes, we use the homography and dominant plane masks of the two images estimated by the homography estimation network and a pre-trained dominant plane detection network, respectively, along with a sampled GT homography to generate a new target image $I^{'}_t$. Then, we use the $I_s$ and $I^{'}_t$, together with the sampled GT homography, to compose a training sample of our supervised dataset. The dominant planes of $I_s$ and $I^{'}_t$ can be fully aligned by the GT homography while the rest remain in realistic motion, making the two images satisfy both label criteria and realism criteria. As shown in Fig.~\ref{fig:teaser}(b), the dominant plane of the generated image pair can be fully aligned by the GT homography, while the scene parallax and dynamic objects are maintained in the foreground.
	
	In addition, few artifacts could exist in the generated target image in early iterations, we, therefore, propose a content consistency module and a quality assessment module to eliminate the unexpected content and select high-quality image pairs for training, respectively. The selected image pairs are used to update the homography estimation network which is utilized to estimate homographies for image generation in the next iteration. During the iterative learning steps, the capability of the network is gradually improved, as well as the quality of the synthesized dataset. With our framework, any unlabeled image pairs can be used to generate training samples, thus addressing the lack of qualified datasets in supervised homography learning, which improves the performance of supervised methods and enables them to be better generalized to real-world scenes. In summary, our main contributions are threefold:
	\begin{itemize}
		\item We propose an iterative deep framework to generate a realistic dataset from unlabeled real-world image pairs for supervised homography learning and simultaneously obtain a high-precision network based on the generated dataset.
		\item We propose a content consistency module and a quality assessment module, achieving the elimination of unexpected content and the selection of qualified data for training.
		\item Experimental results show that our method brings noticeable image realism improvement compared to the prior dataset generation strategy and achieves state-of-the-art performance on public benchmarks.
	\end{itemize}
	
	\section{Related Work}
	\label{sec:related_work}
	\subsection{Traditional Method}
	\label{subsec:traditional methods}
	Traditional homography estimation methods can be divided into feature-based methods and optimization-based methods. Feature-based methods usually combine feature extraction and matching algorithms~\cite{sift,surf,sosnet,superpoint,superglue,loftr} with outliner suppression approaches~\cite{ransac,magsac}, followed by solving DLT~\cite{DLT} to obtain homography. However, such methods would crash in challenging scenes where sufficient feature matches cannot be obtained. Optimization-based methods~\cite{Clkn,DPCP} use the Lucas-Kanade algorithm~\cite{lucas} or calculate the sum of squared differences between two images to optimize an initialized homography iteratively. But optimization-based approaches are time-consuming and suffer from cumulative errors.
	
	\subsection{Learning-based Method}
	\label{subsec:learning-based methods}
	Following the development of deep image alignment methods, such as optical flow~\cite{GyroFlow,ASFlow,RealFlow,KPFlow,AGRFlow} and dense correspondence~\cite{PDC-Net,GLU-Net}, the first deep homography estimation network was proposed by~\cite{supervised2016}. Deep learning-based methods can be divided into two categories: supervised and unsupervised. The former ones~\cite{Dynamic-supervised2020,Crossresolution-supervised2021,Iterative-supervised2022} utilize synthetic image pairs that are derived from single images for training, which suffer from lacking dynamic objects and realistic scene parallax, hindering their generalization ability to real-world scenes. Unsupervised methods~\cite{unsupervised2018,unsupervised2020,CA-Unsupervised2020,BasesHomo2021,HomoGAN2022} are more robust than supervised ones thanks to their label-free training strategies. However, existing unsupervised methods are mostly optimized by minimizing the photometric distance from the warped source to the target, being sensitive to dynamic objects and homogeneous regions~\cite{GLU-Net}. Besides, unsupervised methods are unstable and difficult to converge during the training phase compared to supervised ones. Therefore, our work aims to generate a realistic dataset with homography labels to optimize the network in a supervised manner and enable the network to better generalize to unseen scenes.
	\begin{figure*}[!t]
		\centering
		\includegraphics[width=\linewidth]{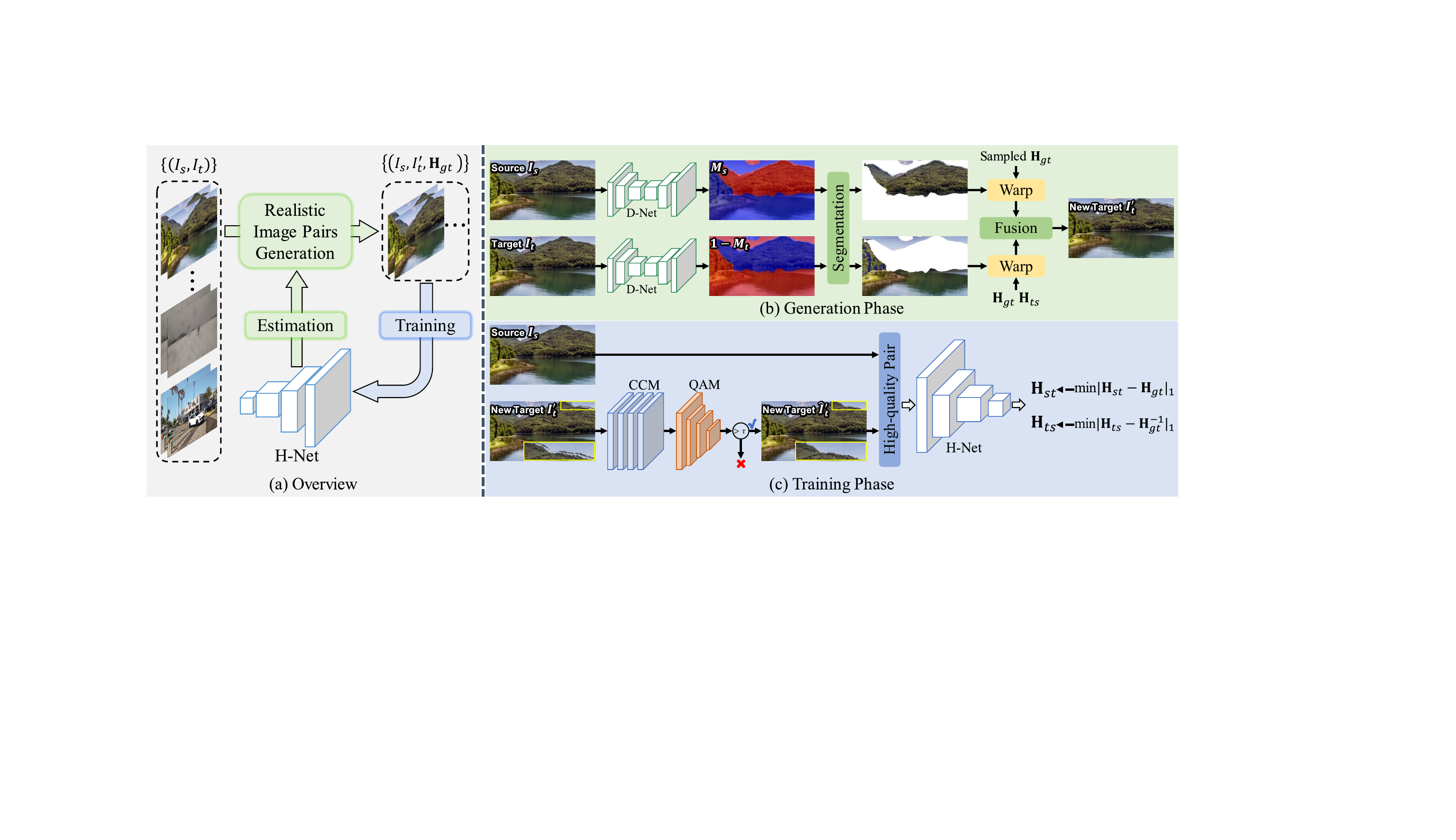}
		\caption{The overall pipeline of our proposed framework. In the generation phase, we synthesize a new target image and form a training sample with the source image according to the dominant plane masks and homography of the original image pair estimated by D-Net and H-Net. In the training phase, we propose a content consistency module and a quality assessment module to prepare qualified image pairs for training H-Net that is adopted for image generation in the next iteration. Moreover, we perform the same pre-processing on the training pairs as in the previous method~\cite{supervised2016,CA-Unsupervised2020,BasesHomo2021,HomoGAN2022}, including central cropping, grayscale transformation, and normalization.}
		\label{fig:Pipeline}
	\end{figure*}
	
	\subsection{Deep Homography Estimation Dataset}
	\label{subsec:Previous Dataset}
	The MS-COCO~\cite{MSCOCO} and CA-unsup~\cite{CA-Unsupervised2020} datasets are commonly used for supervised and unsupervised homography learning. Recently, the GHOF dataset~\cite{gyroflow+} is introduced for unsupervised gyroscope-guided optical flow and homography estimation. The MS-COCO dataset only contains source images, the target images are obtained by warping source images using pre-defined GT homographies. As such, the synthetic image pairs satisfy the label criteria but lack realism criteria. The CA-unsup and GHOF datasets collect image pairs from consecutive real-world video frames that satisfy the realism criteria but lack GT labels. In contrast, our method can synthesize realistic image pairs with labels from any unlabeled image pairs, instead of single images, to satisfy both label and realism criteria.
	
	\section{Method}
	\label{sec:method}
	\subsection{Overview}
	\label{subsec:Overview}
	The overall pipeline of our method is illustrated in Fig.~\ref{fig:Pipeline}. The core innovation of our method is that generating training data and estimating homography are mutually reinforcing. We, therefore, integrate dataset generation and network training into an iterative process, as shown in Fig.~\ref{fig:Pipeline}(a). Our method consists of two phases: the generation phase (G-phase) and the training phase (T-phase).
	
	$\bullet$ \textbf{G-phase}: Given an unlabeled source image $I_s$ and target image $I_t$, we generate a new target image $I^{'}_{t}$ according to a pre-defined homography $\mathbf{H}_{gt}$ as well as the homography and dominant plane masks of the two images estimated by the homography estimation network (H-Net) denoted as $\Theta_{H}$ and the dominant plane detection network (D-Net) as $\Theta_{D}$, forming a training sample as $x = (I_s, I^{'}_{t}, \mathbf{H}_{gt})$, i.e.,
	\begin{equation}
		\label{eqeation:eq1}
		x = \operatorname{G}(I_s,I_t,\Theta_{D},\Theta_{H}).
	\end{equation}
	
	$\bullet$ \textbf{T-phase}: Using the generated training samples $X=\{x\}$ to update the H-Net with the help of a content consistency module (CCM) denoted as $\Theta_{c}$ and a quality assessment module (QAM) as $\Theta_{q}$, i.e,
	\begin{equation}\label{eqeation:eq2}
		\Theta_{H}^{'} = \underset{\theta}{\arg \min} \mathcal{L}\left(X,\Theta_{H},\Theta_{c},\Theta_{q}\right),
	\end{equation}
	where $\Theta_{H}^{'}$ denotes the H-Net retrained on the training samples and $\mathcal{L}$ is the learning objective.
	
	Performing G-phase and T-phase interactively can yield a qualified dataset and a high-precision network.
	
	\subsection{Generation Phase}
	\label{subsec:Generation Phase}
	As shown in Fig.~\ref{fig:Pipeline}(b), we aim to generate a new target image $I^{'}_{t}$ according to the source image $I_s$ and target image $I_t$ along with a sampled homography $\mathbf{H}_{gt}$ to serves as the label between $I_s$ and $I^{'}_{t}$, which is modeled by randomly selecting scaling, shearing, rotation, translation, and perspective factors from pre-defined small-baseline ranges. As a result, a training sample that satisfies both label and realism criteria~\cite{RealFlow} is generated as $(I_s,I^{'}_{t},\mathbf{H}_{gt})$. 
	
\begin{figure}[!t]
		\centering
		\includegraphics[width=\linewidth]{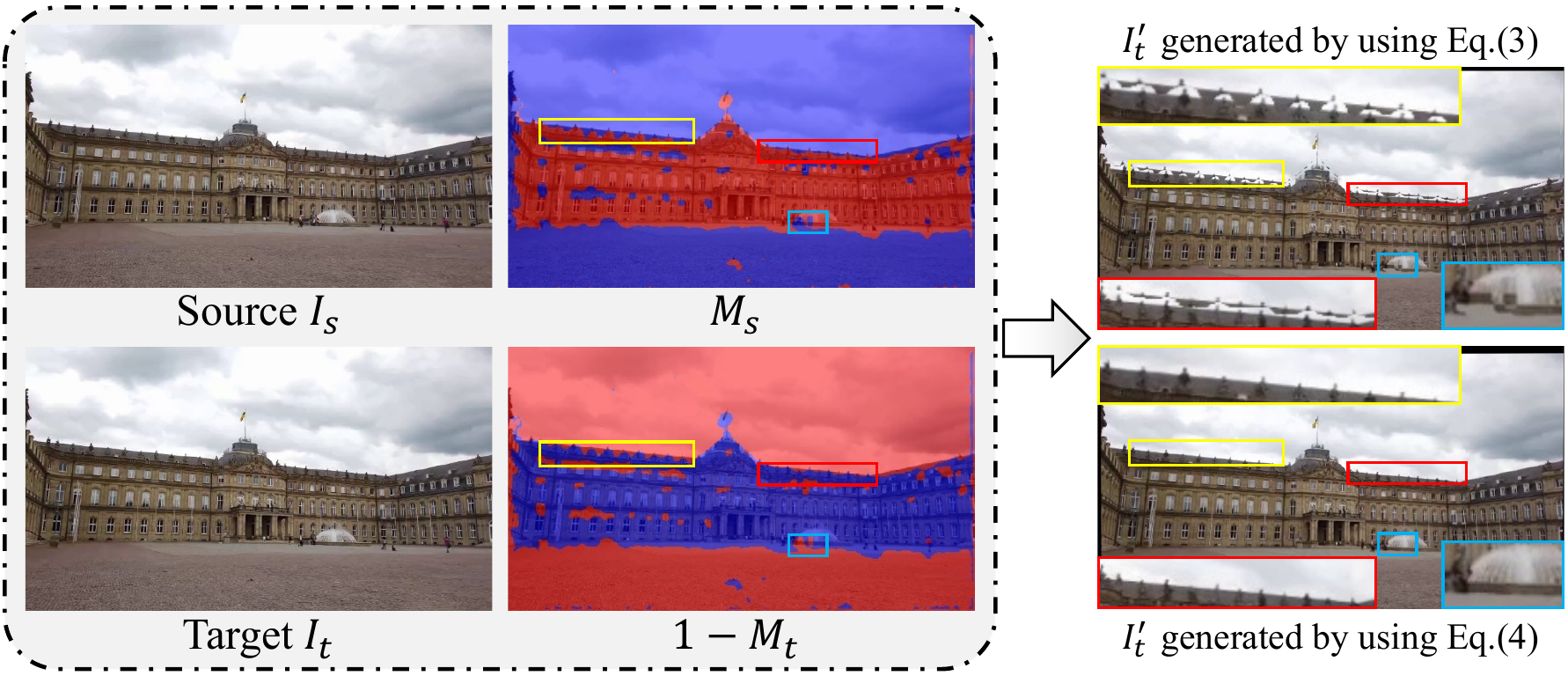}
		\caption{Illustration of generated images by using different fusion functions. The image fused by using Eq.(\ref{eqeation:eq3}) leads to unsmooth edges and artifacts while using Eq.(\ref{eqeation:eq4}) renders a natural image.}
		\label{fig:generation_compare}
	\end{figure}

	Specifically, we first adopt a dominant plane detection network (D-Net), which is formed as a UNet-like structure and pre-trained in an unsupervised manner, to estimate the dominant plane masks $M_s$ and $M_t$ of the $I_s$ and $I_t$, and segment the dominant plane region $P_{d}^{s}=I_s \cdot M_s$ and non-dominant plane region $P_{n}^{t}=I_t \cdot (1 - M_t)$. By warping the $P_{d}^{s}$ with $\mathbf{H}_{gt}$ to form the counterpart region of $I^{'}_{t}$, we ensure that their dominant planes are fully aligned. To model the realistic motion between two images, we construct the non-dominant plane region of $I^{'}_{t}$ from $I_t$. One simple approach is to directly fuse the warped $P_{d}^{s}$ and $P_{n}^{t}$ as 
	\begin{equation}
		\label{eqeation:eq3}
		I^{'}_{t} = \mathcal{W}(P_{d}^{s} + P_{n}^{t}, \mathbf{H}_{gt}),
	\end{equation}
	where $\mathcal{W}(\cdot)$ represents the warp operation. However, there exists relative motion between the two images, and the masks may not be accurate and dense enough to segment continuous and complete regions, artlessly using Eq.(\ref{eqeation:eq3}) to form the new target image would produce discontinuities in the fusion boundary and artifacts. To solve this problem, we construct the non-dominant plane region of $I^{'}_{t}$ by warping the counterpart of $I_t$ with the cumulative multiplication result of the homography from target to source $\mathbf{H}_{ts}$ and $\mathbf{H}_{gt}$, and the final $I^{'}_{t}$ can be obtained as 
	\begin{equation}
		\label{eqeation:eq4}
		I^{'}_{t} = \mathcal{W}(P_{d}^{s}, \mathbf{H}_{gt}) + \mathcal{W}(P_{n}^{t}, \mathbf{H}_{gt} \mathbf{H}_{ts}).
	\end{equation}
	As shown in Fig.~\ref{fig:generation_compare}, despite the inaccuracy of $M_s$ and $M_t$, artifacts are avoided and image smoothness is guaranteed using Eq.(\ref{eqeation:eq4}). 
	
	\subsection{Training Phase}
	\label{subsec:Training Phase}
	Through the G-phase, we obtain a large-scale realistic dataset for improving the homography estimation network (H-Net). As mentioned above, the quality of generated images depends on the accuracy of the estimated dominant plane masks and homography, few artifacts could exist in generated images in early iterations. To this end, we propose a content consistency module and a quality assessment module to refine the generated image and further select qualified image pairs for training, as shown in Fig.~\ref{fig:Pipeline}(c).
	
\begin{figure}[!t]
		\centering
		\includegraphics[width=\linewidth]{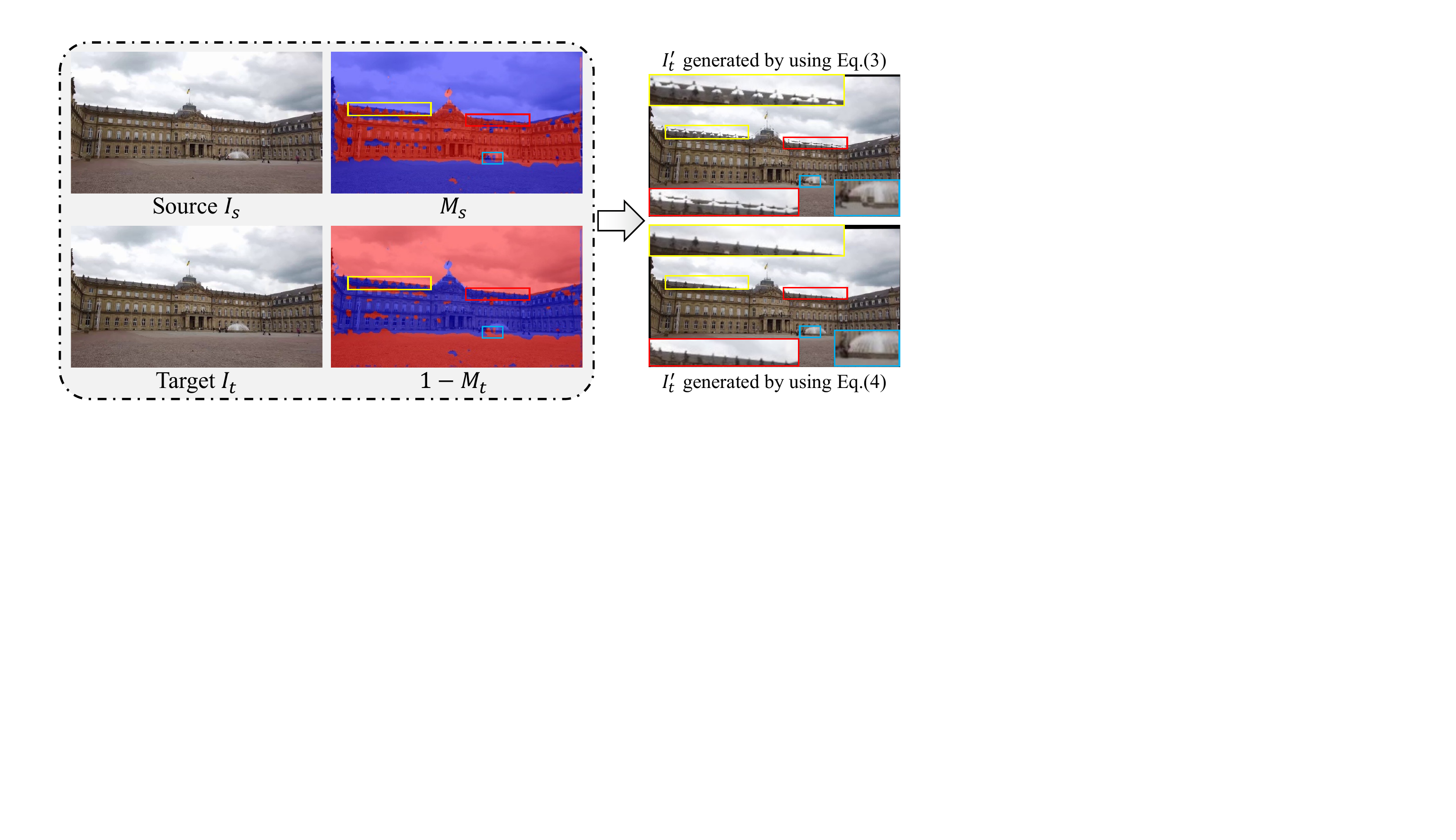}
		\caption{Illustration of the capabilities of our proposed modules. (a) shows that the artifacts are successfully removed by applying our CCM, yielding high-quality images for training. (b) shows the quality scores of different images estimated by QAM, where high-quality images are used for training and inversely rejected.}
		\label{fig:Inpainting_compare}
	\end{figure}

	\textbf{Content Consistency Module.} The content consistency module (CCM) is designed to eliminate artifacts in $I^{'}_{t}$ to improve the content quality. As described in Sec.~\ref{subsec:Generation Phase}, the dominant plane of the $I_s$ and $I^{'}_{t}$ can be fully aligned by the $\mathbf{H}_{gt}$ as $P_{d}^{t^{'}} = \mathcal{W}(P_{d}^{s}, \mathbf{H}_{gt})$, and the $P_{d}^{t}$ and $P_{d}^{s}$ can be aligned by the $\mathbf{H}_{ts}$ as $P_{d}^{s} = \mathcal{W}(P_{d}^{t}, \mathbf{H}_{ts})$ once the $\mathbf{H}_{ts}$ is accurate. Therefore, the dominant plane of $I^{'}_{t}$ can be converted into the counterpart of $I_t$ as
	\begin{equation}
		\label{eqeation:eq5}
		P_{d}^{t^{'}} = \mathcal{W}(\mathcal{W}(P_{d}^{t}, \mathbf{H}_{ts}), \mathbf{H}_{gt}) = \mathcal{W}(P_{d}^{t}, \mathbf{H}_{gt}\mathbf{H}_{ts}).
	\end{equation}
	Besides, the non-dominant plane of $I^{'}_{t}$ is obtained by warping the non-dominant plane of $I_t$ with the accumulated homography $\mathbf{H}_{gt}\mathbf{H}_{ts}$. Therefore, where $\mathbf{H}_{ts}$ is extremely accurate, the $I^{'}_{t}$ can be obtained by warping $I_t$ with the accumulated homography as $I^{'}_{t} = \mathcal{W}(I_t, \mathbf{H}_{gt}\mathbf{H}_{ts})$,  which is the assumption behind the CCM we designed. To achieve this, a content consistency loss $\mathcal{L}_{ccl}$ is proposed as 
	\begin{equation}
		\label{eqeation:eq6}
		\mathcal{L}_{ccl} = |\mathcal{W}(\mathcal{F}(\hat{I}^{'}_{t}), (\mathbf{H}_{gt}\mathbf{H}_{ts})^{-1})-\mathcal{F}(I_t)|_1,
	\end{equation}
	where $\hat{I}^{'}_{t}$ is the reconstructed result of $I^{'}_{t}$ by CCM, i.e., $\hat{I}^{'}_{t}=\Theta_{c}(I^{'}_{t})$, and $\mathcal{F}(\cdot)$ is the feature extrator of H-Net. Instead of directly minimizing the content distance between $\hat{I}^{'}_{t}$ and $I_t$, we constrain the similarity of the features to prompt CCM to reconstruct $\hat{I}^{'}_{t}$ to be sharp as the real-world image, since the $\mathbf{H}_{ts}$ is not accurate enough. As shown in Fig.~\ref{fig:Inpainting_compare}(a), the artifacts in $I^{'}_{t}$ have been successfully removed by CCM. 
	
	\textbf{Quality Assessment Module.} The quality assessment module (QAM) is designed to filter bad cases that cannot be reconstructed by CCM, further improving the quality of the training data. Specifically, the QAM is designed to be a classification network whose output $x \in \mathbb{R}$ ranges from 0 to 1, $x$ being closer to 1 means better quality and vice versa.
	
	To achieve this, we sample another homography to interfere with $\mathbf{H}_{ts}$ since the quality of the generated image is partially relied on the accuracy of $\mathbf{H}_{ts}$, producing a disturbing image $I_r$ with artifacts through G-phase. The score $x_r$ obtained by feeding $I_r$ into QAM should be close to 0. In contrast, the original target image $I_t$ is captured from real-world scenes, so the score $x_t$ of $I_t$ should be close to 1. Thus, the quality assessment loss $\mathcal{L}_{qal}$ is proposed to optimize the QAM as 
	\begin{equation}
		\label{eqeation:eq7}
		\mathcal{L}_{qal} = \operatorname{BCE}(x_t, 1) + \operatorname{BCE}(x_r, 0),
	\end{equation}
	where $\operatorname{BCE}(\cdot)$ denotes the binary cross-entropy loss. As a result, when the score $\hat{x}^{'}_{t}$ of $\hat{I}^{'}_{t}$ exceeds a certain threshold $\tau$, we consider $\hat{I}^{'}_{t}$ to be a high-quality image that can be used for training and inversely be rejected, as shown in Fig.~\ref{fig:Inpainting_compare}(b).
	
	Overall, with the help of CCM and QAM, qualified training pairs can be generated, satisfying both realism criteria and supervision labels to improve the H-Net. Through iterative learning, the performance of the network and the quality of the generated dataset are mutually improved.
	
	\subsection{Network Training}
	\label{subsec:Network Training}
	Since the training samples have GT labels, we can optimize the H-Net by minimizing the difference between $\mathbf{H}_{gt}$ and the estimated $\mathbf{H}_{st^{'}}$. However, as mentioned in~\cite{supervised2016}, it is non-trivial to directly estimate a tomography, we follow previous works~\cite{Dynamic-supervised2020,Iterative-supervised2022,Crossresolution-supervised2021} to use the 4 corners offset vectors $D_{\mathbf{H}_{st^{'}}}$ and $D_{\mathbf{H}_{gt}}$ as supervision objectives, the corresponding homography could be computed by solving DLT~\cite{DLT} using the offset vectors. In addition, we compute the bidirectional homography as the $\mathbf{H}_{t^{'}s}$ should also be aligned with $\mathbf{H}_{gt}^{-1}$. The finally supervised loss $\mathcal{L}_{sup}$ is expressed as 
	\begin{equation}
		\label{eqeation:eq8}
		\mathcal{L}_{sup} = |D_{\mathbf{H}_{st^{'}}} - D_{\mathbf{H}_{gt}}|_1 + |D_{\mathbf{H}_{t^{'}s}} - D_{\mathbf{H}_{gt}^{-1}}|_1.
	\end{equation}
	The total loss $\mathcal{L}_{total}$ is expressed by combing the supervised loss, the content consistency loss, and the quality assessment loss as 
	\begin{equation}
		\label{eqeation:eq9}
		\mathcal{L}_{total} = \mathcal{L}_{sup} + \lambda_1 \mathcal{L}_{ccl} + \lambda_2 \mathcal{L}_{qal},
	\end{equation}
	where $\lambda_1$ and $\lambda_2$ are empirically set as 0.5 and 0.1.
	
	\section{Experiment}
	\label{sec:experiment}
	\begin{table*}
		\begin{center}
			\resizebox{0.97\linewidth}{!}{
				\begin{tabular}{rl|lllllll}
					\toprule
					1) & Methods & \multicolumn{1}{c}{AVG} & \multicolumn{1}{c}{RE} & \multicolumn{1}{c}{LT} & \multicolumn{1}{c}{LL} & \multicolumn{1}{c}{SF} & \multicolumn{1}{c}{LF} \\
					\midrule
					2) & $\mathcal{I}_{3\times3}$ & 6.70 (+1617.95\%) & 7.75 (+3422.73\%) & 7.65 (+1765.85\%) & 7.21 (+1164.91\%) & 7.53 (+1611.36\%) & 3.39 (+993.55\%) \\
					\midrule
					3) & SIFT~\cite{sift}+RANSAC~\cite{ransac} & 1.41 (+261.54\%) & 0.30 (+36.36\%) & 1.34 (+226.83\%) & 4.03 (+607.02\%) & 0.81 (+84.09\%) & 0.57 (+83.87\%) \\
					4) & SIFT~\cite{sift}+MAGSAC~\cite{magsac} & 1.34 (+243.59\%) & 0.31 (+40.91\%) & 1.72 (+319.51\%) & 3.39 (+494.74\%)  & 0.80 (+81.82\%) & 0.47 (51.61\%) \\
					5) & ORB~\cite{orb}+RANSAC~\cite{ransac} & 1.48 (+279.49\%) & 0.85 (+286.36\%) & 2.59 (+531.71\%) & 1.67 (+192.98\%) & 1.10 (+150.00\%) & 1.24 (+300.00\%) \\
					6) & ORB~\cite{orb}+MAGSAC~\cite{magsac} & 1.69 (+333.33\%) & 0.97 (+340.91\%) & 3.34 (+714.63\%) & 1.58 (+177.19\%) & 1.15 (+161.36\%) & 1.40 (+351.61\%) \\
					7) & SPSG~\cite{superpoint,superglue}+RANSAC~\cite{ransac} & 0.71 (+82.05\%) & 0.41 (+86.36\%) & 0.87 (+112.20\%) & 0.72 (+26.32\%) & 0.80 (+81.81\%) & 0.75 (+141.94\%) \\
					8) & SPSG~\cite{superpoint,superglue}+MAGSAC~\cite{magsac} & 0.63 (+61.54\%) & 0.36 (+63.64\%) & 0.79 (+92.68\%) & 0.70 (+22.81\%) & 0.71 (+61.36\%) & 0.70 (+125.81\%) \\
					9) & LoFTR~\cite{loftr}+RANSAC~\cite{ransac} & 1.44 (+269.23\%) & 0.56 (+154.55\%) & 2.70 (+558.54\%) & 1.36 (+138.60\%) & 1.05 (+138.64\%) & 1.52 (+390.32\%) \\
					10) & LoFTR~\cite{loftr}+MAGSAC~\cite{magsac} & 1.39 (+256.41\%) & 0.55 (+150.00\%) & 2.57 (+526.83\%) & 1.33 (+133.33\%) & 1.05 (+138.64\%) & 1.41 (+354.84\%) \\
					\midrule
					11) & CAHomo~\cite{CA-Unsupervised2020} & 0.88 (+125.64\%) & 0.73 (+231.82\%) & 1.01 (+146.34\%) & 1.03 (+80.70\%) & 0.92 (+109.09\%) & 0.70 (+125.81\%) \\
					12) & BasesHomo~\cite{BasesHomo2021} & 0.50 (+28.21\%) & 0.29 (+31.82\%) & 0.54 (+31.71\%) & 0.65 (+14.04\%) & 0.61 (+38.64\%) & 0.41 (+32.26\%) \\
					13) & HomoGAN~\cite{HomoGAN2022} & \underline{0.39 (+0.00\%)} & \underline{0.22 (+0.00\%)} & \underline{0.41 (+0.00\%)} & \underline{0.57 (+0.00\%)} & \underline{0.44 (+0.00\%)} & \underline{0.31 (+0.00\%)}\\
					\midrule
					14) & DHN~\cite{supervised2016} & 2.87 (+635.90\%) & 1.51 (+586.36\%) & 4.48 (+992.68\%) & 2.76 (+384.21\%) & 2.62 (+495.45\%) & 3.00 (+867.74\%) \\
					15) & LocalTrans~\cite{Crossresolution-supervised2021} & 4.21 (+978.26\%) & 4.09  (+1757.59\%) & 4.84  (+1081.14\%) & 4.55  (+697.98\%) & 5.30  (+1105.20\%) & 2.25  (+624.29\%) \\
					16) & IHN~\cite{Iterative-supervised2022} & 4.67 (+1097.44\%) & 4.85 (+2104.55\%) & 5.54 (+1251.22\%) & 5.10 (+794.74\%) & 5.04 (+1045.45\%) & 2.84 (+816.13\%)\\
					\midrule
					17) & $\textrm{DHN}^*$~\cite{supervised2016} & 1.89 (+384.95\%) & 1.21 (+451.82\%) & 2.13 (+418.54\%) & 2.33 (+307.91\%) & 1.72 (+291.23\%) & 2.07 (+567.81\%) \\
					18) & $\textrm{LocalTrans}^*$~\cite{Crossresolution-supervised2021} & 1.76 (+352.03\%) & 1.25 (+467.41\%) & 2.15 (+423.85\%) & 1.90 (+232.61\%) & 2.25 (+410.41\%) & 1.28 (+311.77\%) \\
					19) & $\textrm{IHN}^*$~\cite{Iterative-supervised2022} & 1.19 (+205.54\%) & 0.72 (+227.09\%) & 1.74 (+324.93\%) & 1.18 (+107.42\%) & 1.30 (+196.52\%) & 1.01 (+225.58\%) \\
					20) & Ours & \textbf{0.34 (-12.82\%)} & \textbf{0.22 (+0.00\%)} & \textbf{0.35 (-14.63\%)} & \textbf{0.44 (-22.81\%)} & \textbf{0.42 (-4.55\%)} & \textbf{0.29 (-6.45\%)} \\
					\bottomrule
			\end{tabular}}
		\end{center}
		\caption{The point matching errors (PME) of our method and all comparison methods on the CA-unsup~\cite{CA-Unsupervised2020} test set. The best and second-best results are highlighted in \textbf{bold} and \underline{underlined}. The percentages in the parentheses indicate the relative change in comparison to the second-best results. SPSG indicates SuperPoint with SuperGlue, and $*$ denotes the methods are retrained on our CA-sup dataset.}
		\label{tab:Compare_CAunsup}
	\end{table*}
	
	\subsection{Dataset}
	\label{subsec:Dataset}
	\textbf{MS-COCO.} The MS-COCO dataset~\cite{MSCOCO} is one of the widely used datasets in the fields of object detection~\cite{detection}, segmentation~\cite{Segmentation}, etc. Previously supervised methods~\cite{supervised2016,Dynamic-supervised2020,Crossresolution-supervised2021,Iterative-supervised2022} used it to prepare their training datasets by perturbing each corner of single images to produce GT homographies and warp the images using the homographies to generate training pairs.
	
	\textbf{CA-Unsup.} The CA-unsup dataset~\cite{CA-Unsupervised2020} contains 800k training pairs and 4.2k testing pairs that are captured from five types of real-world scenes, i.e., regular (RE), low texture (LT), low light (LL), small foreground (SF), and large foreground (LF), where the last four are challenging scenes for homography estimation. The training set contains only unlabeled image pairs, but for each pair of the test set, 6$\thicksim$10 equally distributed matching points are manually marked for evaluation. Based on the CA-unsup dataset, we generated a new dataset that contains the same number of image pairs satisfying both label criteria and realism criteria as the CA-unusp dataset through our proposed method, named \textbf{CA-sup}.  We show some examples of our CA-sup dataset in the supplementary materials.
	
	\textbf{GHOF.} The GHOF dataset~\cite{gyroflow+} is designed to evaluate homography and optical flow estimation with the gyroscope readings, it contains 10k training data and 300 testing pairs under 5 different categories, including regular (RE), foggy (FOG), low light (LL), rainy (RAIN), and snowy (SNOW) scenes. Compared to the CA-unsup dataset, it contains more parallax motion variations and extreme foreground ratio which are challenging to the homography methods. The sparse 5$\thicksim$8 correspondences are used to evaluate the results.
	
	\subsection{Implementation Details}
	\label{subsec:Implementation}
	Our framework consists of D-Net, CCM, QAM, and H-Net. The D-Net is pre-trained on the CA-unsup dataset in an unsupervised manner, the detailed architecture and implementation details are described in the supplementary material. We choose BasesHomo~\cite{BasesHomo2021}, which balances performance and efficiency as the backbone of H-Net, and convert its output representation into 4 corners offset vectors. The CCM and QAM are training with the H-Net for 2 iterations, and the training parameters are consistent with the official implementation of BasesHomo for each iteration.
	
	\subsection{Comparison with Existing Methods}
	\label{subsec:Comparison}
	\textbf{Comparison Methods.} We compare our method with three categories of existing homography estimation methods: 1) feature-based methods including SIFT~\cite{sift}, ORB~\cite{orb}, SuperPoint~\cite{superpoint} with SuperGlue~\cite{superglue} (SPSG), and LoFTR~\cite{loftr}, 2) supervised methods including DHN~\cite{supervised2016}, LocalTrans~\cite{Crossresolution-supervised2021}, and IHN~\cite{Iterative-supervised2022}, 3) unsupervised methods including CAHomo~\cite{CA-Unsupervised2020}, BasesHomo~\cite{BasesHomo2021}, and HomoGAN~\cite{HomoGAN2022}. For feature-based methods, we improve them with two different outlier rejection algorithms RANSAC~\cite{ransac} and MAGSAC~\cite{magsac}, respectively. For supervised methods, we employ their models pre-trained on the MS-COCO dataset~\cite{MSCOCO} and the models retrained on our CA-sup dataset denoted with $*$ for evaluation.
	
	\textbf{Quantitative Comparison.} We report the quantitative results of all comparison methods on the CA-unsup test set in Table~\ref{tab:Compare_CAunsup}. The $\mathcal{I}_{3\times3}$ in the first row refers to a 3 × 3 identity matrix as a “no-warping” homography for reference, of which the errors reflect the original distance between point pairs. As shown in Table~\ref{tab:Compare_CAunsup}, our method achieves state-of-the-art performance in all categories and outperforms the best existing method HomoGAN by 12.82\%, with the points matching error (PME) reduced from 0.39 to 0.34. The feature-based methods can perform well in regular (RE) scenes, as sufficient matching points can be obtained, while our methods and HomoGAN reduce the error by 0.08 compared to SIFT+RANSAC. More specifically, our method produces a PME of 0.216 which is lower than 0.222 of HomoGAN. The small foreground (SF) and large foreground (LF) scenes contain dynamic objects that affect the accuracy of estimated homography, and our method has the lowest PME in these two categories, even when compared to the methods utilizing outlier removal masks for robust estimation, i.e., CAHomo and HomoGAN. In low texture (LT) and low light (LL) scenes, most learning-based methods are more robust due to their keypoint-free estimation strategy, especially unsupervised ones. However, homogeneous region~\cite{GLU-Net} dominates a large portion of the image in the LT and LL scenes, where the photometric loss between the two images is minor, so the performance of unsupervised methods in these two categories is not as well as in others. In contrast, our method outperforms the unsupervised methods in the LT and LL with errors reduced by at least 14.63\% and 22.81\% respectively, thanks to our photometric-free fully supervised learning strategy.
	\begin{table}[!t]
		\begin{center}
			\resizebox{0.98\linewidth}{!}{
				\begin{tabular}{rl|ccccccc}
					\toprule
					1) & Methods & AVG & RE & FOG & LL & RAIN & SNOW \\
					\midrule
					2) & $\mathcal{I}_{3\times3}$ & 6.33  & 4.94  & 7.24  & 8.09  & 5.48  & 5.89  \\
					\midrule
					3) & SIFT~\cite{sift}+MAGSAC~\cite{magsac} & 3.90 & \textbf{0.64} & 4.01 & 9.77 & 0.70 & 4.40 \\
					4) & ORB~\cite{orb}+RANSAC~\cite{ransac} & 15.14 & 6.92 & 31.27 & 27.80 & 1.82 & 7.90 \\
					5) & SPSG~\cite{superpoint,superglue}+MAGSAC~\cite{magsac} & 3.28 & 3.41 & 1.46 & 7.61 & 0.75 & 3.19 \\
					6) & LoFTR~\cite{loftr}+MAGSAC~\cite{magsac} & 2.55 & 2.76 & 1.16 & 5.34 & 0.52 & 2.98  \\
					\midrule
					7) & CAHomo~\cite{CA-Unsupervised2020} & 3.87  & 4.10  & 3.84 & 6.99  & 1.27  & 3.17  \\
					8) & BasesHomo~\cite{BasesHomo2021} & 2.28  & 2.02  & 1.43  & 4.90  & 0.78  & 2.29  \\
					9) & HomoGAN~\cite{HomoGAN2022} & \underline{1.95} & 1.73 & \textbf{0.60} & \textbf{3.95} & \underline{0.47} & 3.02 \\
					\midrule
					10) & DHN~\cite{supervised2016} & 6.61  & 6.04  & 6.02  & 7.68  & 6.99  & 6.32  \\
					11) & LocalTrans~\cite{Crossresolution-supervised2021} & 5.72 & 4.06 & 6.49 & 5.95 & 5.78 & 6.34 \\
					12) & IHN~\cite{Iterative-supervised2022} & 8.17 & 7.10 & 8.71 & 9.34 & 6.57 & 9.13 \\
					\midrule
					13) & $\textrm{DHN}^*$~\cite{supervised2016} & 3.01 & 1.92 & 3.94 & 4.54 & 1.98 & 2.66 \\
					14) & $\textrm{LocalTrans}^*$~\cite{Crossresolution-supervised2021} & 2.89 & 1.78 & 4.27 & 4.59 & 1.37 & 2.43 \\
					15) & $\textrm{IHN}^*$~\cite{Iterative-supervised2022} & 2.59 & 2.21 & 3.05 & 4.70 & 0.98 & \underline{2.03} \\ 
					16) & Ours & \textbf{1.72} & \underline{1.60} & \underline{0.88} & \underline{4.42} & \textbf{0.43} & \textbf{1.28} \\
					\bottomrule
			\end{tabular}}
		\end{center}
		\caption{The point matching errors (PME) of our method and comparison methods on the GHOF~\cite{gyroflow+} test set. The best and second-best results are highlighted in \textbf{bold} and \underline{underlined}.}
		\label{tab:Compare_GHOF}
	\end{table}%
	
	\begin{figure*}[!ht]
		\centering
		\includegraphics[width=0.96\linewidth]{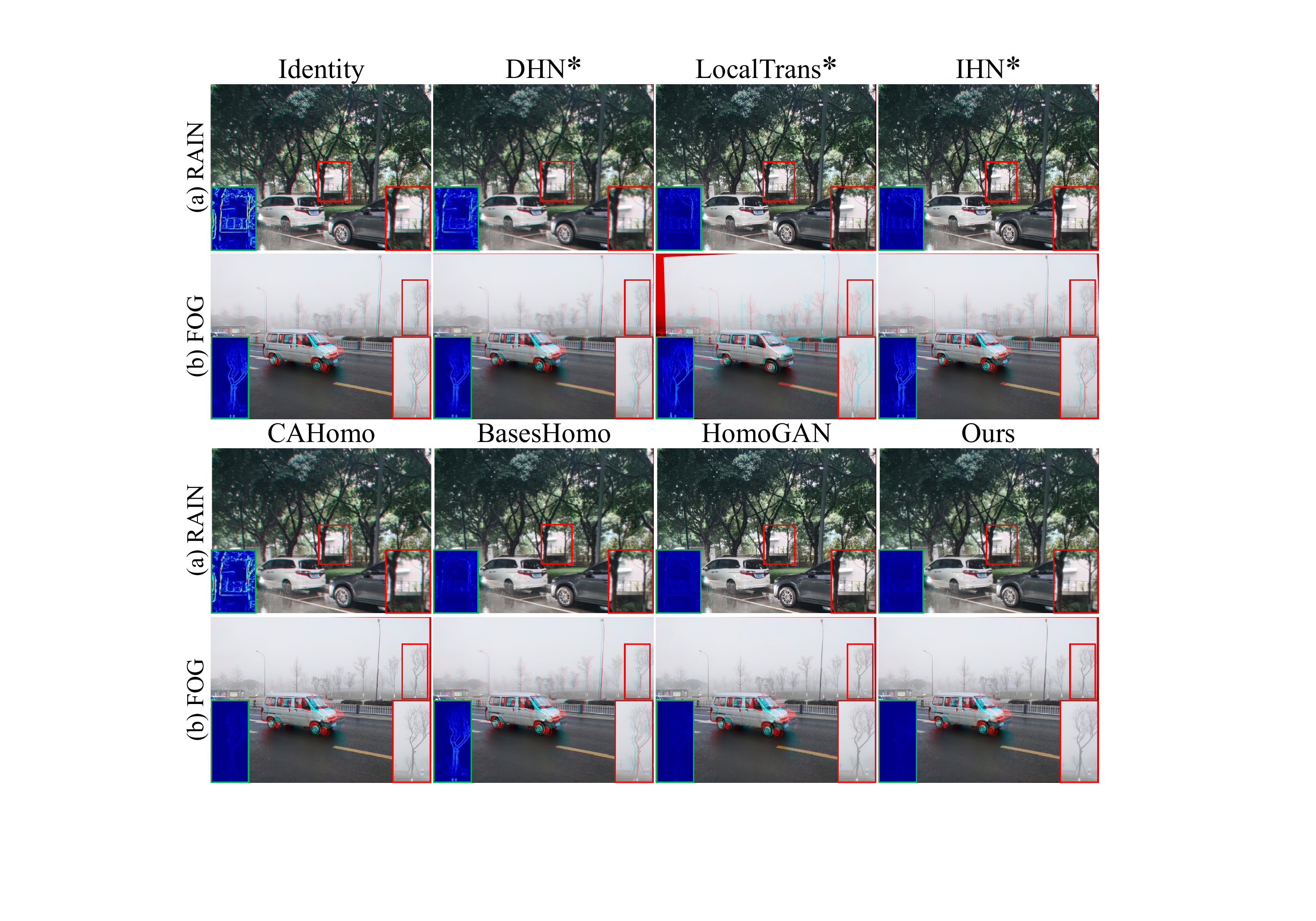}
		\caption{Qualitative results of our method and other competitive methods on the CA-unsup~\cite{CA-Unsupervised2020} test set. The images are generated by superimposing the warped source images on the target image. Error-prone regions are highlighted with red boxes, and the green boxes show the content difference between the two images in the error-prone regions. Best viewed by zooming in.}
		\label{fig:CA_unusp visual}
	\end{figure*}
	\begin{figure}[!t]
		\centering
		\includegraphics[width=0.98\linewidth]{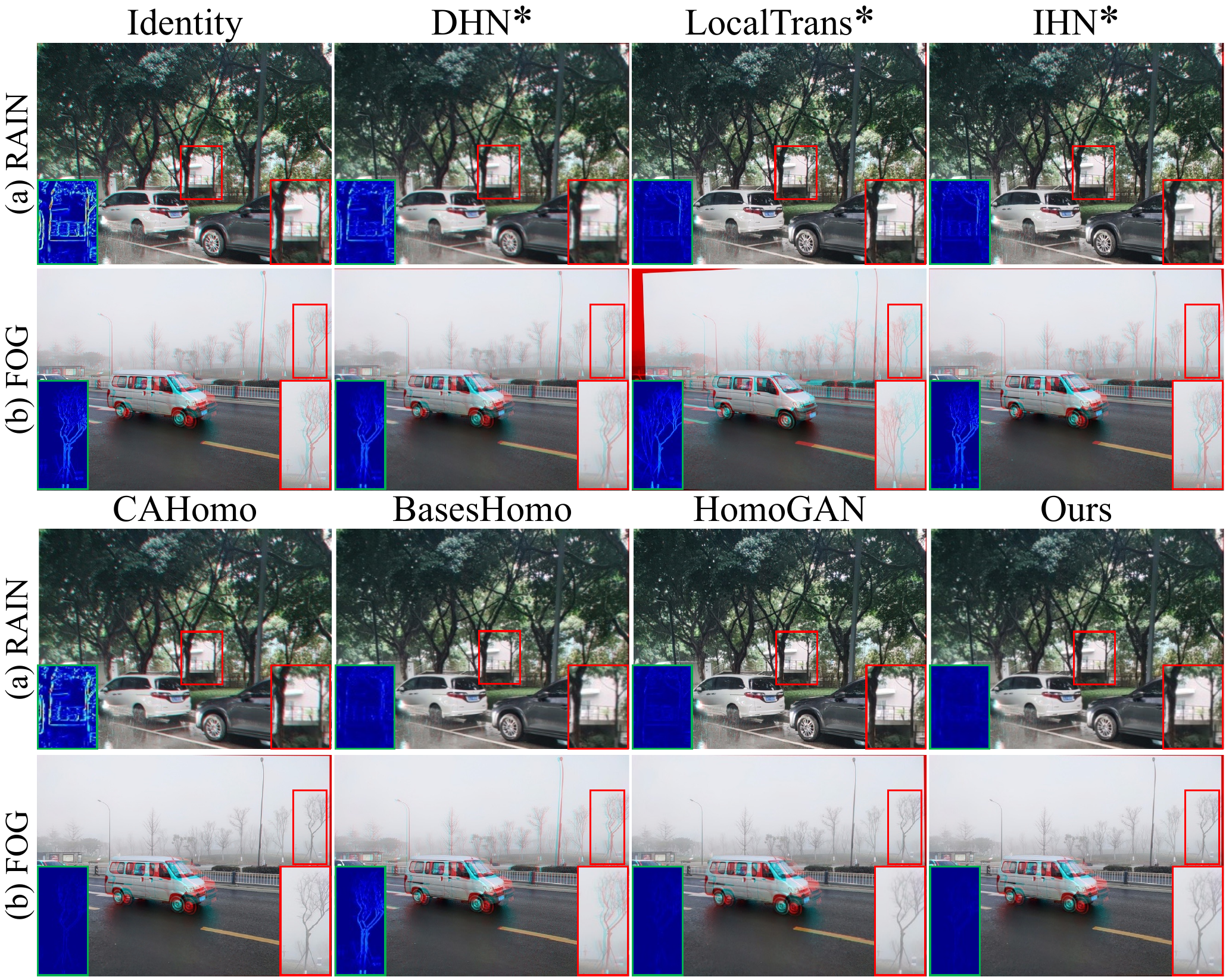}
		\caption{Qualitative results of our method and other competitive methods on the GHOF~\cite{gyroflow+} test set. Best viewed by zooming in.}
		\label{fig:GHOF visual}
	\end{figure}
	
	Moreover, the generalization ability of supervised methods has been widely criticized. Therefore, we further evaluate feature-based methods that performed well on the CA-unsup dataset, learning-based methods, and our method on the GHOF test set to prove the effectiveness of our method. As shown in Table~\ref{tab:Compare_GHOF}, our method achieves the best results in RAIN and SNOW scenes and the second-best results in RE, FOG and LL scenes, resulting in the lowest average PME. For regular scenes, even though feature-based methods can extract sufficient high-quality feature matches, we still outperform most of them. In addition, our CA-sup dataset does not contain images captured in the FOG and RAIN scenes, while our method is capable of generalizing these scenes, which proves the superiority of our method. On the other hand, previous supervised methods cannot generalize well to real-world scenes since their training data lack realistic motion. After retraining on our CA-sup dataset, their performance has been significantly improved, which further proves the effectiveness of our framework.
	
	\textbf{Qualitative Comparison.} In Fig.~\ref{fig:CA_unusp visual} and Fig.~\ref{fig:GHOF visual}, we provide the qualitative results of our method and competitive methods on the CA-unsup and GHOF test sets, respectively. Fig.~\ref{fig:CA_unusp visual}(a) and (b) are from LF and SF scenes with dynamic objects, and our method produces more accurate results than feature-based and unsupervised methods which are sensitive to moving objects, as highlighted in the red and green boxes. Fig.~\ref{fig:CA_unusp visual}(c) is from the LT scene, where feature-based methods all fail due to the lack of sufficient correspondence. On the contrary, learning-based methods perform well, but they are still not competitive as ours. In Fig.~\ref{fig:GHOF visual}, we provide the comparison results in RAIN and FOG scenes of the GHOF test set, where our training set does not contain images captured in such scenes. As mentioned above, while unsupervised methods have better generalization ability in unseen scenes, we prove that supervised methods can also achieve satisfactory results after training on our generated dataset. For example, in Fig.~\ref{fig:GHOF visual}(a), the previous supervised methods can even obtain more accurate results than CAHomo, and our method performs the best. 
	
	\textbf{Comparison with Dataset Generation Methods.} We use images from the CA-unsup dataset to generate a new dataset named $\textrm{CA-sup}^\dagger$ using the previous dataset generation strategy~\cite{supervised2016}, and retrain H-Net on it for comparison. Quantitative results on the CA-unsup test set are illustrated in Table~\ref{tab:Dataset generation strategy}. The previous strategy generates image pairs from single images that meet label criteria only, causing unsatisfactory performance. In contrast, the image pairs contained in our CA-sup dataset can help the network to learn real-world motion representations, proving the importance of realistic scene motion for estimating homography.
	\begin{table}[!t]
		\begin{center}
			\resizebox{0.96\linewidth}{!}{
				\begin{tabular}{l|c|cccccc}
					\toprule
					& Dataset & AVG & RE & LT & LL & SF & LF \\
					\midrule
					DHN~\cite{supervised2016} &$\textrm{CA-sup}^\dagger$ & 1.37 & 0.79 & 1.70 & 1.28 & 1.23 & 1.83  \\
					\midrule
					Ours & CA-sup & 0.34 & 0.22 & 0.35 & 0.44 & 0.42 & 0.29 \\
					\bottomrule
			\end{tabular}}
		\end{center}
		\caption{Comparison with previous dataset generation method. We use the same source images to generate the training dataset and train the same network for comparison.}
		\label{tab:Dataset generation strategy}
	\end{table}
	
\begin{figure}[!t]
		\centering
		\includegraphics[width=\linewidth]{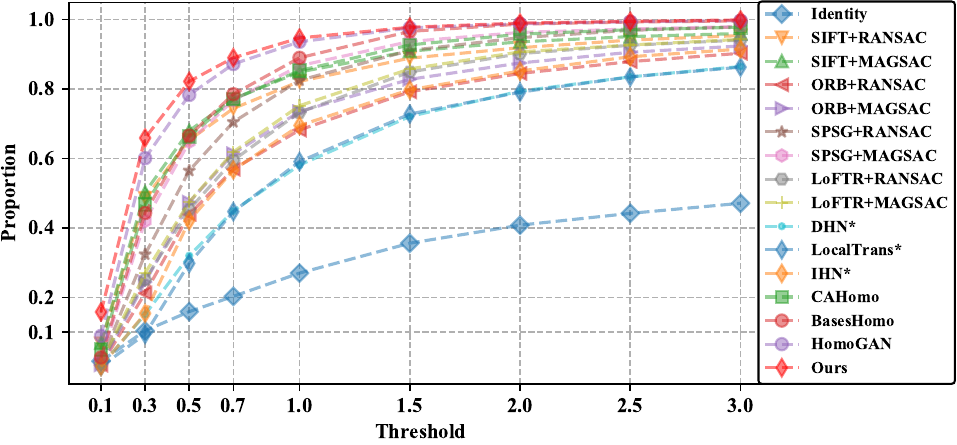}
		\caption{The proportion of inliers of our method and all comparison methods under various thresholds. The higher position of curves represents better robustness.}
		\label{fig:robustness}
	\end{figure}

	\textbf{Robustness Evaluation.}
	Furthermore, we evaluate the robustness of all comparison methods on the CA-unsup test set by setting thresholds to calculate the proportion of inlier predictions. As shown in Fig.~\ref{fig:robustness}, we plot a series of curves where axis $X \in [0.1, 3.0]$ is the threshold and axis $Y \in [0.0, 1.0]$ is the percentage of inliers, higher position of curves represents better robustness. With a threshold of 0.1, our outlier percentage is 6.9\% higher than the second-best method (15.8\% vs. 8.9\%), and our method achieves a proportion of 99.9\% with a threshold of 3.0.
	
	\subsection{Ablation Study}
	\label{subsec:ablation}
	In this section, we conduct a series of ablation studies to measure the impact of different component choices of our method. We use the CA-sup dataset for training, and quantitative results on the CA-unsup test set are illustrated in Table~\ref{tab:ablation_studies}, detailed experiment settings are discussed below. For more ablation experiments, please refer to the supplementary material.
	
	\textbf{Dataset Generation Strategy.} We conduct an experiment to evaluate the effectiveness of our dataset generation strategy by using G-phase only to generate training samples to train the network, denoted as Baseline in row 1 of Table~\ref{tab:ablation_studies}. Even without using CCM and QAM, our method achieves state-of-the-art performance compared to the existing methods. Moreover, the homography estimation network, i.e., BasesHomo, trained on our generated dataset achieves a 24\% performance gain compared to the network trained with the unsupervised learning strategy, further proving the superiority of our dataset generation strategy.
	
	\textbf{Module Effectiveness.} We conduct experiments to evaluate the effectiveness of our proposed CCM and QAM by removing them separately from the overall framework. As shown in rows 2-3 of Table~\ref{tab:ablation_studies}, CCM and QAM can help to generate and select high-quality image pairs for training, achieving error reduction of 0.03 and 0.01, respectively, compared to the Baseline in row 1 using none of them. By using both CCM and QAM in the training phase, our method can achieve the best performance.
	
	\textbf{Different Homography Estimation Networks.} As described in Sec.~\ref{subsec:Implementation}, we choose BasesHomo as the backbone of our H-Net. In rows 4-5 of Table~\ref{tab:ablation_studies}, we further switch the backbone to CAHomo and HomoGAN, denoted as Real-CA and Real-GAN, respectively, to prove that our method also works not only on a specific architecture. Compared with the unsupervised results in rows 11 and 13 of Table~\ref{tab:Compare_CAunsup}, CAHomo obtains performance gain by 50\% with our framework and HomoGAN reduces the error by 0.09, proving the generalization of our proposed framework.
	
\begin{figure}[!t]
		\centering
		\includegraphics[width=\linewidth]{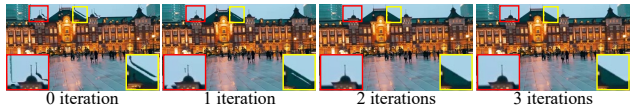}
		\caption{Illustration of the generated image in different iterations.}
		\label{fig:iteration_compare}
	\end{figure}

	\textbf{Iteration Times.} The quality of generated dataset and network performance are mutually reinforcing through iterative learning. As shown in rows 6-8 of Table~\ref{tab:ablation_studies}, the performance of H-Net improves with iteration increases, while higher-precision H-Net generates more natural images, as shown in Fig.~\ref{fig:iteration_compare}, and high-quality images can improve performance. When the homography estimated by H-Net is accurate enough, there would be no artifacts and boundary discontinuities in the generated images. Notably, ``0 iteration" indicates that the image is generated by using the homography estimated by unsupervised BasesHomo. However, a certain upper limit exists in our method and it will converge after several iterations. The performance gain of 3 iterations compared to 2 iterations, i.e., default, is not significant, and more iterations are not cost-effective compared to the resources spent per iteration.
	\begin{table}[!t]
		\begin{center}
			\resizebox{0.98\linewidth}{!}{
				\begin{tabular}{rl|cccccc}
					\toprule
					& & AVG & RE & LT & LL & SF & LF \\
					\midrule
					1) & Baseline & 0.38 & 0.23  & 0.40  & 0.50  & 0.45  & 0.33  \\
					\midrule
					2) & w/o CCM & 0.37 & 0.23 & 0.36 & 0.48 & 0.44 & 0.34 \\
					3) & w/o QAM & 0.35  & 0.23  & 0.36  & 0.45  & 0.42  & 0.30  \\
					\midrule
					4) & Real-CA~\cite{CA-Unsupervised2020} & 0.44 & 0.27 & 0.45 & 0.52 & 0.53 & 0.43 \\
					5) & Real-GAN~\cite{HomoGAN2022} & 0.30 & 0.18 & 0.30 & 0.39 & 0.37 & 0.28 \\
					\midrule
					6) & 1 iteration & 0.38 & 0.24 & 0.40 & 0.46 & 0.48 & 0.33 \\
					7) & 3 iterations & 0.33 & 0.21 & 0.34 & 0.44 & 0.40 & 0.28 \\
					\midrule
					8) & Default & 0.34 & 0.22 & 0.35 & 0.44 & 0.42 & 0.29 \\
					\bottomrule
			\end{tabular}}
		\end{center}
		\caption{Results of ablation studies, please refer to the text for more details. w/o denotes without.}
		\label{tab:ablation_studies}
	\end{table}
	
	\section{Conclusion}
	\label{sec:conclusion}
	We have presented an iterative deep framework that aims to generate a realistic dataset for supervised homography learning and obtain a high-precision network. We use the dominant plane masks estimated by D-Net and the homography estimated by H-Net to generate training pairs, and the CCM and QAM are proposed to prepare high-quality training data to update the H-Net. The trained network and the generated data can be improved iteratively, yielding a qualified dataset as well as a state-of-the-art network. Extensive experiments demonstrate the superiority of our method and the effectiveness of our newly proposed components, the performance of existing methods can be also improved using our dataset for training.
	
	\textbf{Acknowledgements.} This work is supported by Sichuan Science and Technology Program under grant No.2023NSFSC0462.
	
	{\small
		\bibliographystyle{ieee_fullname}
		\bibliography{egbib}
	}
	
\end{document}